\documentclass[10pt]{article}
\usepackage[a4paper,left=0.79in,top=0.98in,textwidth=6.69in,textheight=9.92in]{geometry}
\usepackage{mathptmx}
\author{Ilias Giannakopoulos}
\usepackage{graphicx}
\usepackage[caption=false]{subfig}
\usepackage{mathrsfs,amsmath,amsfonts,amssymb,amsthm,amstext,amscd,amsxtra,amsopn}
\usepackage{diagbox}
\usepackage{lettrine}
\usepackage{cite}
\usepackage{eucal}
\usepackage{soul}

\usepackage{bm}
\usepackage{multirow}
\usepackage{esint}
\usepackage{empheq}
\usepackage{comment}
\usepackage{boxhandler}
\usepackage{bm}
\usepackage{algorithm}
\usepackage[noend]{algpseudocode}
\usepackage{etoolbox}
\apptocmd{\sloppy}{\hbadness 10000\relax}{}{}
\usepackage{leftidx}
\usepackage{makecell}
\usepackage{stackrel}
\usepackage{xcolor}
\usepackage{hyphenat}
\usepackage{nicefrac}
\usepackage[normalem]{ulem}
\usepackage[hyphens]{url}
\usepackage[hidelinks]{hyperref}
\usepackage{fancyhdr}
\hypersetup{breaklinks=true}
\begin{document}
\begin{center}
\textbf{\large{MR-Based Electrical Property Reconstruction Using Physics-Informed Neural Networks}} \\[0.2cm]
\small{Xinling Yu}$^1$, \small{Jos\'{e} E. C. Serrall\'{e}s}$^2$, \small{Ilias I. Giannakopoulos}$^3$, \small{Ziyue Liu}$^4$, \small{Luca Daniel}$^2$, \small{Riccardo Lattanzi}$^3$, \small{Zheng Zhang}$^1$ \\[0.2cm]
$^1$\textit{\small{Department of ECE, University of California at Santa Barbara, Santa Barbara, CA, USA;}}
$^2$\textit{\small{Department of EECS, Massachusetts Institute of Technology, Cambridge, MA, USA;}}
$^3$\textit{\small{Center for Advanced Imaging Innovation and Research (CAI2R) and the Bernard and Irene Schwartz Center for Biomedical Imaging, Department of Radiology, New York University Grossman School of Medicine, New York, NY, USA;}}
$^4$\textit{\small{Department of Statistics and Applied Probability, University of California at Santa Barbara, Santa Barbara, CA, USA}}
\\[0.6cm]
\end{center}

\noindent
\textbf{Introduction:} Electrical properties (EP), namely permittivity and electric conductivity, dictate the interactions between electromagnetic waves and biological tissue [1]. EP can be potential biomarkers for pathology characterization, such as cancer, and improve therapeutic modalities, such radiofrequency hyperthermia and ablation. MR-based electrical properties tomography (MR-EPT) uses MR measurements to reconstruct the EP maps. Using the homogeneous Helmholtz equation, EP can be directly computed through calculations of second order spatial derivatives of the measured magnetic transmit or receive fields $(B_{1}^{+}, B_{1}^{-})$. However, the numerical approximation of derivatives leads to noise amplifications in the measurements and thus erroneous reconstructions [3]. Recently, a noise-robust supervised learning-based method (DL-EPT) was introduced for EP reconstruction [4]. However, the pattern-matching nature of such network does not allow it to generalize for new samples since the network's training is done on a limited number of simulated data. In this work, we leverage recent developments on physics-informed deep learning [5] to solve the Helmholtz equation for the EP reconstruction. We develop deep neural network (NN) algorithms that are constrained by the Helmholtz equation to effectively de-noise the $B_{1}^{+}$ measurements and reconstruct EP directly at an arbitrarily high spatial resolution without requiring any known $B_{1}^{+}$ and EP distribution pairs.
\\

\noindent
\textbf{Methods:} Assuming a smooth distribution of the EP, the relationship between the transmit fields and the underlying EP can be described by the Helmholtz equation: $\nabla^{2}B_{1}^{+} + k_{0}^{2}\varepsilon_{c}B_{1}^{+} = 0,$ where $k_{0}$ is the wave number in free space and $\varepsilon_{c}$ is the relative complex permittivity. We aim to de-noise the measured complex transmit field $\{(\bm{r_{i}}, \tilde{B}_{1}^{+}(\bm{r_{i}}))\}_{i=1}^{N}$ at $N$ 3D locations $\bm{r_i}$ and reconstruct the EP. To do so, we employ prior physical knowledge as a constraint in processing the noisy measurements. We define a NN $\mathcal{B}_{1}^{+}(\bm{r};\bm{\theta_{1}})$, parameterized by a set of weights and biases $\bm{\theta_{1}}$, to estimate the complex $B_{1}^{+}$ fields. We use an additional NN $\mathcal{E}_{c}(\bm{r};\bm{\theta_{2}})$ with independent weights and biases $\bm{\theta_{2}}$ to estimate the distribution of relative complex permittivity. The Helmholtz residual then takes the form
\begin{equation}
\mathcal{R}_{H} = \nabla^{2}\mathcal{B}_{1}^{+}(\bm{r};\bm{\theta_{1}}) + k_{0}^{2}\mathcal{E}_{c}(\bm{r};\bm{\theta_{2}})\mathcal{B}_{1}^{+}(\bm{r};\bm{\theta_{1}}).
\end{equation}
Here the Laplacian of the NN representation can be readily computed to machine precision using automatic differentiation [6]. A good set of candidate parameters $\{\boldsymbol{\theta_{1}, \theta_{2}}\}$ can be obtained via gradient descent using a composite loss function that aims to fit the measured $\tilde{B}_{1}^{+}$ field while also penalizing the Helmholtz equation residual
\begin{equation}
\begin{aligned}
\mathcal{L}(\bm{\theta_{1}, \theta_{2}}) &=\mathcal{L}_{\text {data }}(\bm{\theta_{1}})+\lambda \mathcal{L}_{r}(\bm{\theta_{1}, \theta_{2}}), \:\: \text{where} \:\: \mathcal{L}_{r}(\bm{\theta_{1}, \theta_{2}}) = \frac{1}{N}\sum_{i=1}^{N} |\mathcal{R}_{H}(\bm{r}_{i}, \bm{\theta_{1}, \theta_{2}})|^{2}, \:\:\text{and} \\
\mathcal{L}_{\text {data }}(\bm{\theta_{1}}) &=  \frac{1}{N}\sum_{i=1}^{N} |\mathrm{Re}\{\mathcal{B}_{1}^{+}(\bm{r}_{i};\bm{\theta_{1}})\} - \mathrm{Re}\{\tilde{B}_{1}^{+}(\bm{r_{i}})\}|^{2} + |\mathrm{Im}\{\mathcal{B}_{1}^{+}(\bm{r}_{i};\bm{\theta_{1}})\} - \mathrm{Im}\{\tilde{B}_{1}^{+}(\bm{r_{i}})\}|^{2}.
\label{eq3}
\end{aligned}
\end{equation}
For all experiments, $\mathcal{B}_{1}^{+}$ and $\mathcal{E}_{c}$ were two independent, fully-connected NN with 3 layers, 128 units per hidden layer, and a sine activation function. We train these NN jointly by minimizing the loss of equation \eqref{eq3} using the Adam optimizer for 120k iterations in total, with a decaying schedule of learning rates $10^{-3}, 10^{-4}, 10^{-5}$, decreased every 40k iterations.\\

\noindent
\textbf{Results and Discussion:} We consider a 4-compartment phantom, presented in Figure 1 a) with permittivity $\epsilon_{r} =\{56,51,$ $65,76\}$ and electric conductivity $\sigma =\{0.69,0.56,0.84,1.02\}$. We use one external excitation to generate synthetic transmit fields maps inside the phantom at 7 tesla MRI frequency. We set $\lambda = 10^{-6}$ in \eqref{eq3} to make each component of the loss approximately equal at the beginning of training. Once trained, the resulting physics-informed NN $\mathcal{B}_{1}^{+}(\bm{r};\bm{\theta_{1}})$ and $\mathcal{E}_{c}(\bm{r};\bm{\theta_{2}})$ can be used to obtain physically consistent predictions of $B_{1}^{+}$ and EP at any arbitrary 3D location. Three noisy $\tilde{B}_{1}^{+}$ maps are shown in Figure~\ref{fig1} b). We present the de-noised $B_{1}^{+}$ maps and the reconstructed EP for the central axial cut of the phantom in Figure~\ref{fig1} c), d) and e). The average value of the peak normalized absolute error (PNAE) over the entire volume is $0.23\%, 3.52\%$ and $4.45\%$ for the $B_{1}^{+}$, relative permittivity and conductivity, respectively, when peak SNR = 100; The error is found to be $0.44\%, 2.55\%$ and $3.43\%$ for the $B_{1}^{+}$, relative permittivity and conductivity, respectively, when peak SNR = 20. These results show that our algorithm is accurate and robust for a significant amount of noise. Our algorithm can also reconstruct the EP using incomplete measurements. For example, when half of the $\tilde{B}_{1}^{+}$ measurements of the entire volume are missing, as in Figure 1 b) right, we can still reconstruct the $B_{1}^{+}$ and the EP in the whole domain of interest, in which case the error is $0.24\%, 2.49\%$ and $3.68\%$ for the $B_{1}^{+}$, relative permittivity and conductivity, respectively. \\

\renewcommand{\thefigure}{1}
\begin{figure}[ht!]
\begin{center}
\includegraphics[width=0.75\textwidth]{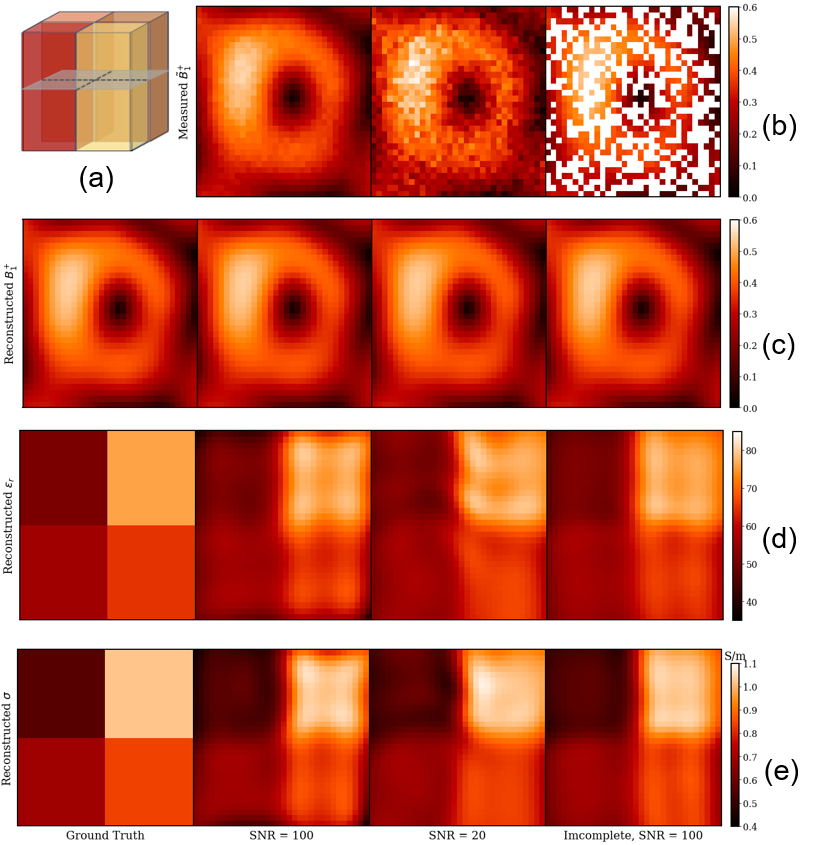}
\caption{(a) 4-compartment phantom. (b) From left to right, $\tilde{B}_{1}^{+}$ measurements (peak SNR = 100, 20), incomplete $\tilde{B}_{1}^{+}$ measurements (peak SNR = 100). (c) Ground truth $B_{1}^{+}$ compared with (from left to right) de-noised $B_{1}^{+}$ measurements (peak SNR = 100, 20), reconstructed $B_{1}^{+}$ for incomplete $B_{1}^{+}$ measurements (SNR = 100). (d) Ground truth $\varepsilon_{r}$ compared with (from left to right) reconstructed $\varepsilon_{r}$ (SNR = 100, 20), reconstructed $\varepsilon_{r}$ using incomplete $B_{1}^{+}$ measurements (SNR 100). (e) Ground truth $\sigma$ compared with (from left to right) reconstructed $\sigma$ (SNR = 100, 20), reconstructed $\sigma$ using incomplete $B_{1}^{+}$ measurements (SNR = 100).}
\label{fig1}
\end{center}
\end{figure}

\noindent
\textbf{Conclusions:} In this work, we have proposed a physics-informed deep learning framework that is able to de-noise simulated $B_{1}^{+}$ measurements and provide accurate EP reconstructions of an inhomogeneous phantom. To our best knowledge, it is the first time that deep learning can reconstruct the $B_{1}^{+}$ from incomplete noisy measurements, which shows the potential to improve other MR-based reconstruction methods. In future work, we will explore the usage of multiple $B_{1}^{+}$ magnitudes and relative phases (measurable in MRI), discarding the assumption of knowing the absolute phase.\\

\noindent
\textbf{\small{References:}} \\[0.1cm]
[1] Hand, J. W. “Modelling the interaction of electromagnetic fields (10 MHz–10 GHz) with the human body: methods and applications.” Physics in Medicine \& Biology 53.16 (2008): R243. \\[-0.1cm]
[2] Katscher, Ulrich, et al. “Determination of electric conductivity and local SAR via B1 mapping.” IEEE transactions on medical imaging 28.9 (2009): 1365-1374. \\[-0.1cm]
[3] Mandija, Stefano, et al. “Error analysis of Helmholtz-based MR-electrical properties tomography.” Magnetic resonance in medicine $80.1$ (2018): 90-100. \\[-0.1cm]
[4] Mandija, Stefano, et al. “Opening a new window on MR-based electrical properties tomography with deep learning.” Scientific reports 9.1 (2019): 1-9. \\[-0.1cm]
[5] Raissi, Maziar, et al. “Physics-informed neural networks: A deep learning framework for solving forward and inverse problems involving nonlinear partial differential equations.” Journal of Computational physics 378 (2019): 686-707. \\[-0.1cm]
[6] Baydin, Atilim Gunes, et al. “Automatic differentiation in machine learning: a survey.” Journal of Machine Learning Research 18 (2018): 1-43. 
\end{document}